\pdfoutput=1
\documentclass[11pt]{article}
\usepackage{verbatim}
\usepackage{graphicx}  
\usepackage{acl}
\usepackage{multirow}
\usepackage{times}
\usepackage{latexsym}
\usepackage{enumerate}
\usepackage{float}
\usepackage[T1]{fontenc}
\usepackage{amsmath}
\usepackage{color}
\usepackage{booktabs}

\usepackage[utf8]{inputenc}

\usepackage{microtype}

\title{Multimodal Neural Machine Translation \\ with Search Engine Based Image Retrieval}

\author{Zhenhao Tang \ \ \ \ \ \    Xiaobing Zhang  \\
College of Application and Technology, \\
Shenzhen University, \\
Shenzhen, China \\
\\\And
Zi Long\thanks{*Corresponding author} \ \ \ \ \ \    Xianghua Fu \\
College of Big Data and Internet, \\
Shenzhen Technology University, \\
Shenzhen, China \\
}
\begin{document}
\maketitle

\begin{abstract}
\label{sec:abs}
Recently, numbers of works shows that the performance of neural machine translation (NMT) can be improved to a certain extent with using visual information.
However, most of these conclusions are drawn from the analysis of experimental results based on a limited set of bilingual sentence-image pairs, such as \emph{Multi30K}.
In these kinds of datasets, the content of one bilingual parallel sentence pair must be well represented by a manually annotated image, which is different with the actual translation situation. 
Some previous works are proposed to addressed the problem by retrieving images from exiting sentence-image pairs with topic model. However, because of the limited collection of sentence-image pairs they used, their image retrieval method is difficult to deal with the out-of-vocabulary words, and can hardly prove that visual information enhance NMT rather than the co-occurrence of images and sentences.
In this paper, we propose an open-vocabulary image retrieval methods to collect descriptive images for bilingual parallel corpus using image search engine. Next, we propose text-aware attentive visual encoder to filter incorrectly collected noise images.
Experiment results on \emph{Multi30K} and other two translation datasets show that our proposed method achieves significant improvements over strong baselines.

\end{abstract}

\section{Introduction}
With the development of NMT, the role of visual information in machine translation has attracted researchers' attention\citep{elliott2017findings2,barrault2018findings3,specia2016shared}.
Although we are still not clear about the specific role of visual information in NMT\citep{caglayan2019probing,elliott2018adversarial}, visual information can assist NMT model to achieve better translation performance \citep{calixto2017incorporating,calixto2017doubly,su2021bi-co}.
Different with those text-only NMT\citep{bahdanau2014attn,gehring2016convolutional}, a bilingual parallel corpora with manual image annotations are used to train a multimodal NMT model by an end-to-end framework, and therefore, most of the previous conclusions are drawn from the analysis of experimental results based on a limited set of manually annotated bilingual sentence-image pairs, specifically, \emph{Multi30K}\citep{elliott2016Multi30K}.

In \emph{Multi30K}, as shown in \ref{tab:dataset}, the sentences consists mostly of common and simple words, and the content of each bilingual parallel sentence pair is well represented by a single image.   
%
Table \ref{tab:dataset} also shows an example of bilingual sentence-image pair from an actual news report of United Nations News\footnote{
\url{https://news.un.org/en/}}. 
It is obviously that there is a dramatic difference between the data of  \emph{Multi30K} and the real-world multimodal translation situations. 
Therefore, results and evidences based on \emph{Multi30K} can hardly proved the effectiveness of multimodal NMT model in an actual translation situation, in which sentences contain rare and uncommon words and are partially described by images. 

\begin{table}
    \centering
    \includegraphics[scale=0.44]{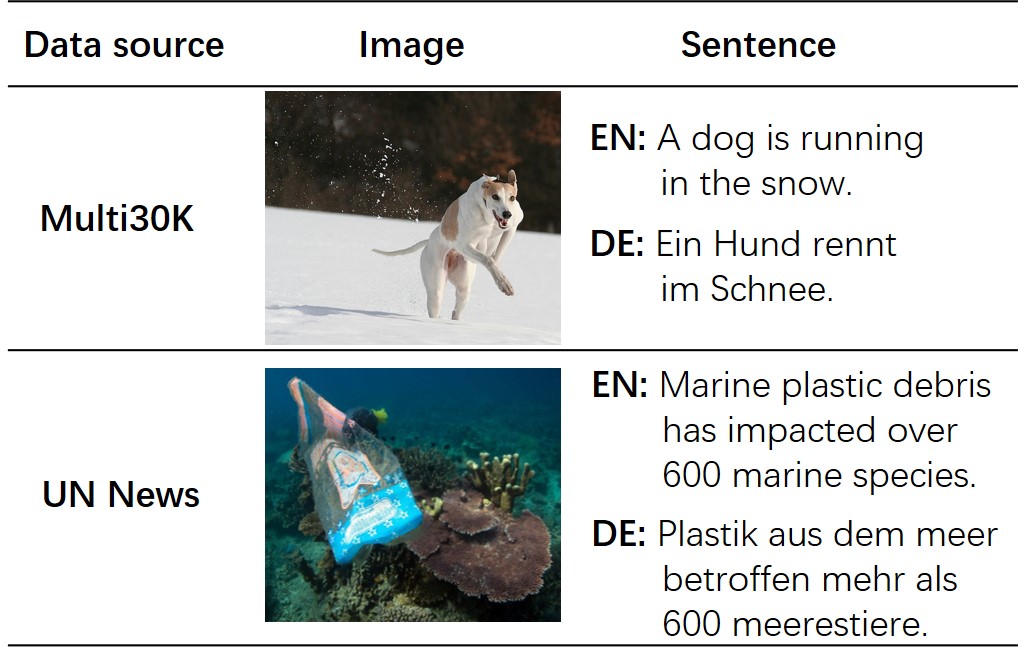}
    \caption{Comparison of example from \emph{Multi30K} dataset and United Nations News.}
    \label{tab:dataset}
\end{table}

%

To address the problem, \citet{zhang2019UVR} proposed to transform the existing sentence-image pairs into a topic-image lookup table, and a group of images with similar topics to the source sentence is retrieved from the topic-image lookup table. 
However, the topic-image lookup table is made from a limited collection of sentence-image pairs, such as \emph{Multi30K} and MS COCO image caption dataset \citep{coco}, their image retrieval method is difficult to deal with the out-of-vocabulary words. Besides, results from  \citet{zhang2019UVR} can hardly prove that the performance of NMT is improved by visual information rather than the co-occurrence of images and sentences. Their model may suffer problems in translating sentences with images that are not contained in the topic-image lookup table.

In this paper, 
we propose an open-vocabulary image retrieval methods to collect images for bilingual parallel corpus using image search engine, thus addressing the problems caused by limited collection of sentence-image pairs in \citet{zhang2019UVR}. 
In detail, to focus on the major part of the sentence, we apply the term frequency-inverse document frequency (TF-IDF).
Instead of a single keyword, we use multiple words as search query for image retrieval to ensure that the contents of collected images are partially consistent with the given sentences.
%
%
Since the quality of images from search engine may be varied, we propose to apply a simple but effective attention layer, and introduce a text-aware attentive visual encoder to filter incorrectly collected noise images.
The proposed method is then evaluated on three translation datasets, including the \emph{Multi30K} English-to-German, WMT'16 English-to-German, Global Voices \citep{GVoice} English-to-German.
Experiment results show that our proposed method achieves significant improvements over strong baselines.
%
To summarize, out contributions are primarily three-fold:
\begin{description}
\item[(1)] We present an open-vocabulary image retrieval methods with image search engine that overcomes the shortcomings of \citet{zhang2019UVR} caused by limited image collection.

\item[(2)] The proposed method enables the text-only NMT to use visual information from the collected images that are partially consistent with input sentences, which is more close to the actual translation situations.

\item[(3)] We further discuss the influence of visual information in the proposed multimodal NMT model , which verified the effectiveness and generality of the proposed approach.

\end{description}

\section{Related Work}
\label{ref:rel}
Recently, multimodal NMT models have gradually become a hot topic in machine translation research. They use image information to improve the translation effect of NMT models through different methods.

In some cases, visual features are directly used as supplementary information to the text presentation. For example, \citet{huang2016attention} takes global visual features and local visual features as additional information for sentences. \citet{calixto2017incorporating} initializes the encoder hidden states or decoder hidden states through global visual features.
\citep{calixto2017doubly} uses an independent attention mechanism to capture visual representations. \citep{caglayan2016multimodal} incorporates spatial visual features into the multimodal NMT model via an independent attention mechanism. On this basis, \citet{delbrouck2017compact} employs Compact Bilinear Pooling to fuse two modalities. \citet{su2021bi-co} introduces image-text mutual interactions to refine their semantic representations. \citet{lin2020dynamic} attempts to
introduce the capsule network into multimodal NMT, they use the timestep-specific source-side context vector to guide the routing procedure.

All the above work is performed on the \emph{Multi30K} dataset.
However, some recent studies indicate that the visual features may play a less important role in the NMT model than previously thought. \citep{ive2019distilling,zhang2017nict,gronroos2018memad}.
Such problems are mainly caused by the limitations of the \emph{Multi30K} dataset.
\citet{zhang2019UVR} presents a universal visual representation method that overcomes the shortcomings of \emph{Multi30K} dataset.
However, all their image information still comes from \emph{Multi30K}, which is obviously not enough to represent complex machine translation corpus.

\section{Background}
\label{section3}
\label{background}
In this section, we give a simple description of the multimodal NMT model proposed by \citet{calixto2017doubly}. The multimodal NMT model is composed of one text encoder, one visual encoder and one decoder with two attention mechanisms. The multimodal NMT aims to construct an end-to-end neural network to model $P = (Y|X,I)$ as follows:
\begin{eqnarray}
\log p(Y|X,I) = \sum\limits_{i=1}^{M}\log p(y_{t}|y_{<t},C,A) \nonumber 
\end{eqnarray}
where $I$ represents visual features, $X = (x_{1},x_{2},\ldots,x_{L})$ is the source sentence, and $Y = (y_{1},y_{2},\ldots,y_{M})$ is the target sentence. The text encoder is a Bi-directional Recurrent Neural Network (RNN) with Gated Unit(GRU)\citep{cho2014GRU} and learn a time-dependent text hidden states $C=(h_{1},h_{2},\ldots,h_{N})$ for the source sentence. The visual encoder is a pretrained convolutional neural network (CNN)  and a visual representation $A$ for the given image.

The decoder is a conditional GRU (cGRU)\footnote{https://github.com/nyu-dl/dl4mt-tutorial/blob/master/docs/cgru.pdf} with two separate attention mechanisms. The text attention mechanism generates a time-dependent context vector $c_{t}$ based on the text hidden states $C$ and the hidden state proposal $s_{t}^{'}$ as follows:
\begin{eqnarray}
c_{t} = f_{att\_text}(C,s_{t}^{'}) \label{1}
\end{eqnarray}
Meanwhile, the visual attention computes a time-dependent context vector $i_{t}$ based on the visual feature maps A and the hidden state proposal $s_{t}^{'}$ as follows:
\begin{eqnarray}
i_{t} = f_{att\_img}(A,s_{t}^{'}) \label{2}
\end{eqnarray}
Where $s_{t}^{'}$ is calculated by the previous hidden state $s_{t-1}$ and the previously generated target word $y_{t-1}$.





\section{Our Proposed Method}
\begin{figure*}[th]
\centering
\includegraphics[scale=0.47]{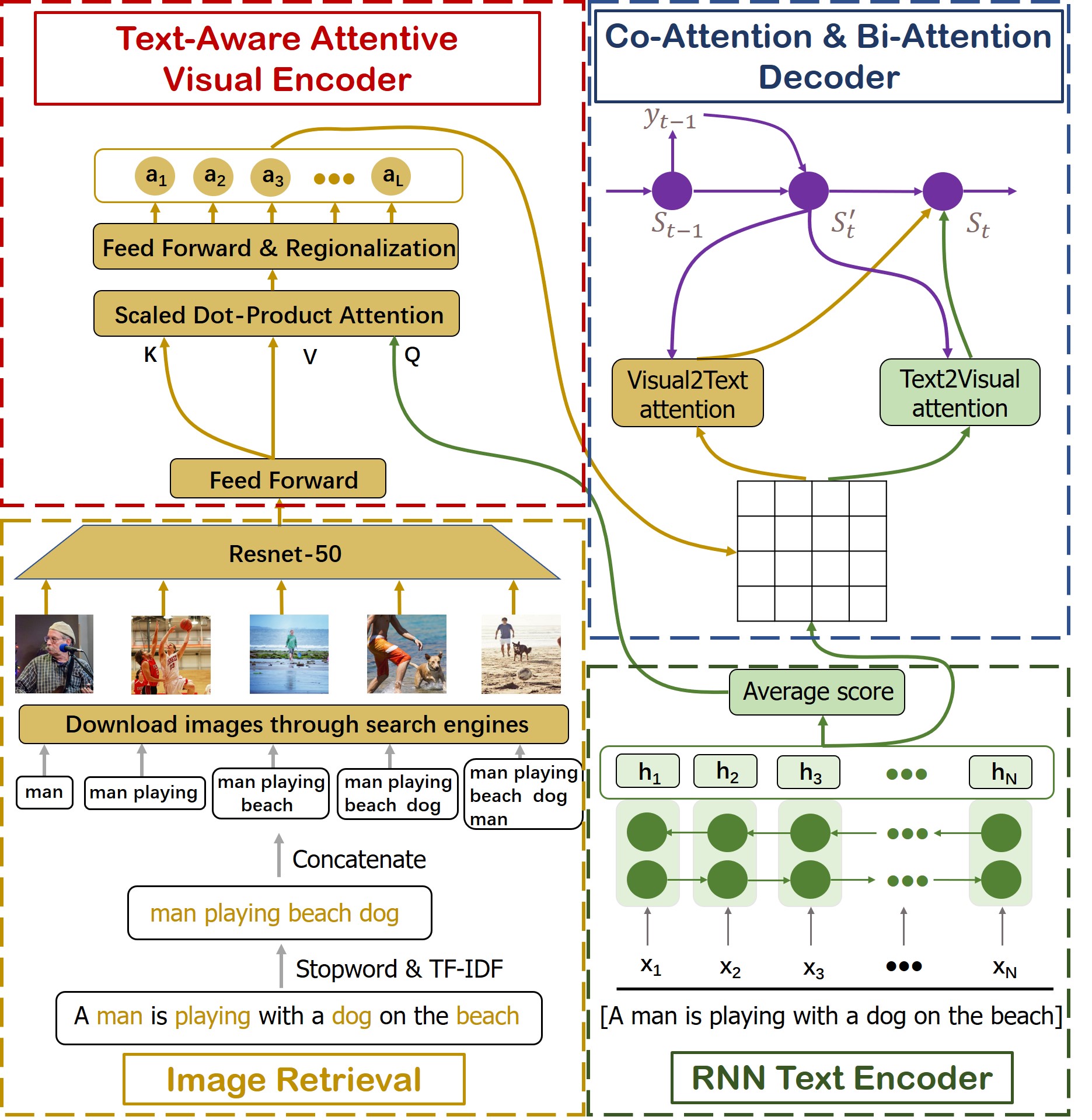}
\caption{The overview of the framework of our proposed method}
\label{fig:model}
\end{figure*}

Figure \ref{fig:model} shows the 4 components of our proposed method, consisting of image retrieval, text-aware attentive visual encoder, RNN text encoder and translation decoder with co-attention \& bi-attention. 

\subsection{Image Retrieval}
\label{image_load}
In this section, we will introduce the proposed open-vocabulary image retrieval methods using image search engine. 

Similar with \citet{zhang2019UVR}, to focus on the major part of the sentence and suppress the noise such as stopwords and low-frequency words, we apply the term frequency-inverse document frequency (TF-IDF) \citep{witten2005kea} to create search queries  for image search engines.
Specifically, given the $i$th ($i=1, 2, \ldots, N$, $N$ represents the number of samples in the training set) source language sentence $X_{i} = \{x_{i}^{1},x_{i}^{2},\ldots,x_{i}^{L}\}$ of length L, $X_{i}$ is first filtered by as stopword list\footnote{
\url{https://github.com/stopwords-iso/stopwords-en}
}, and the filtered input sentence $X_{i}^{f}$ is obtained.  
We then regard $X_{i}^{f}$ as a document $d_{i}$, and compute the TF-IDF score $TI_{i,j}$ for each word $x_{i}^{j}$ ($j=1, 2, \ldots, L$) in $d_{i}$. 
The formula is as follows:
\begin{eqnarray}
TI_{i,j} = \frac{n_{i,j}}{\sum_{k}n_{i,k}} \times \log\frac{|D|}{1+|\{k| x_{i}^{j}\in d_{k}]\}|} \nonumber 
\end{eqnarray}
where $n_{i,j}$ is the number of occurrences of the word $x_{i}^{j}$ in document $d_{i}$, $\sum_{k}n_{i,k}$ represents the total number of words in document $d_{i}$. $|D|=N$ represents the total number of source language sentences in the training data, and $ |\{k|x_{i}^{j}\in d_{k}]\}|$ represents the number of sentences including $x_{i}^{j}$ in the dataset. 
For input sentence $X_{i}$, words are then listed in descending order by $TI_{i,j}$ score, represented as $Q_{i}=(x_{i}^{t_{1}},x_{i}^{t_{2}},\ldots,x_{i}^{t_{L}})$ ($TI_{i,t_{1}} \ge TI_{i,t_{2}} \ge \ldots \ge TI_{i,t_{L}}$). 

Instead of using the top-$k$ high TF-IDF words separately, 
we concatenate several words from the top-$k$ high TF-IDF words as search query.
Specifically, for the sorted words list $Q_{i}$, 
the $m$th search query $q_{m}$ is defined as following:
\begin{eqnarray}
q_{m} = {\rm concat}(x_{i}^{t_{1}}, x_{i}^{t_{2}}, \ldots, x_{i}^{t_{m}}) \nonumber
\end{eqnarray}
Where ${\rm concat}(\cdot)$ means that words are concatenated with blanks as separator.
search query $q_{m}$ is then applied in image search engine and the first available image is collected as the $m$th image for input sentence $X_{i}$, represented as $A_{i}^{m}$.
According to the results of preliminary experiment, we build 5 search queries and collect 5 images for each sentence\footnote{
In the preliminary experiments, we find that the proposed image retrieval method collect less noise and achieves a slightly better translation performance than the method that uses a single word as search query. 
}.

\subsection{Text-Aware Attentive Visual Encoder}
\label{subsec:img_enc}

For each collected image, we employ a 50-layer Residual Network (ResNet-50) \citep{he2016res50} to represent the visual semantic information as a $196 \times 1024$ feature vector.

As described in Section \ref{image_load}, for source language sentence $X_{i}$, we collect 5 images $A_{i}^{1}, A_{i}^{2}, \ldots, A_{i}^{5}$ using image search engine.  In order to filter the incorrectly collected noise images, we apply a simple but effective scaled dot-product attention in visual encoder, where the visual representation $A_{i}$ of input sentence $X_{i}$ is defined as the following formula:
\begin{eqnarray}
A_{i} = \sum^{5}_{m=1} \alpha_{i,m} A_{i}^{m} \nonumber
\end{eqnarray}
where $\alpha_{i,m}$ represents the weight of $m$th images for input sentence $X_{i}$. The $\alpha_{i,m}$ is then computed as follows:
\begin{eqnarray}
 \alpha_{i,m} & = & {\rm softmax}(W (A_{i}^{m}) \cdot {C}_{i}^{'}) \nonumber \\
 {C}_{i}^{'} & = & \frac{1}{N}\sum\limits_{t=1}^{N} h_{i}^{t} \nonumber
\end{eqnarray}
%
where ${\rm softmax}(\cdot)$ stands for softmax activation function, and ${C}_{i}^{'}$ represents an average pool of the hidden states $C_{i}= (h_{i}^{1},h_{i}^{2},\ldots,h_{i}^{N})$ for input sentence $X_{i}$.

Finally, the obtained $196 \times 1024$D visual representation is considered as a matrix  $A_{i} = (\boldsymbol{a}_{i}^{1}, \boldsymbol{a}_{i}^{2},\ldots,\boldsymbol{a}_{i}^{L})$,  $\boldsymbol{a}_{i}^{l}\in R^{1024}$. Each of the $L=196$ rows consists of a 1024D feature vector that represents a specific image region.
Visual representation $A_{i} = (a_{i}^{1},a_{i}^{2},\ldots,a_{i}^{L})$ and text representation $C_{i}= (h_{i}^{1},h_{i}^{2},\ldots,h_{i}^{N})$ are then used as the inputs of tanslation decoder .

\subsection{Translation Decoder}
\label{subsec:dec}

\begin{figure}[ht]
    \centering
    \includegraphics[scale=0.25]{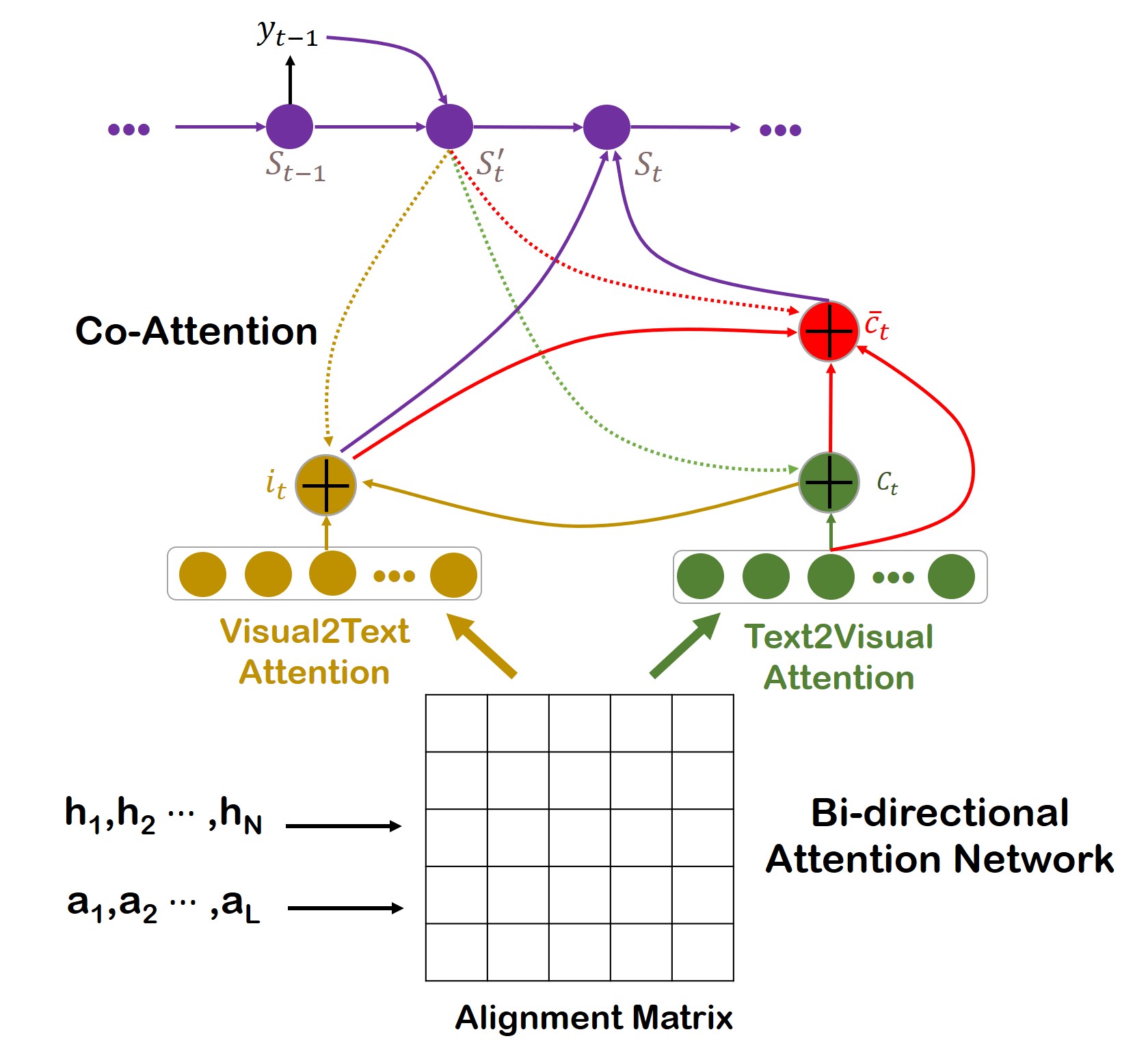}
    \caption{multimodel NMT model with deep semantic interactions}
    \label{fig:bi_model}
\end{figure}
As shown in figure \ref{fig:bi_model}, we apply a bi-directional attention network\footnote{\textcolor{black}{According to the result of the preliminary experiment, we found that Transformer-based model can hardly produce an advantage in performance on such small dataset as \emph{Multi30K}. Therefore, we chosed LSTM as our basic model. As a future work, we are going to integrate Transformer into our proposed method and evaluate it on some larger datasets.}} and a co-attention network \citep{su2021bi-co} to model underlying semantic interactions between text and image. 

The bi-directional attention network is used to enhance text and image representations. Specifically,
we use text representation $C_{i}= (h_{i}^{1},h_{i}^{2},\ldots,h_{i}^{N})$ and visual representation $A_{i} = (a_{i}^{1},a_{i}^{2},\ldots,a_{i}^{L})$ for bi-direction attention network to obtain a shared alignment matrix $S \in R^{N \times L}$.The alignment matrix is computed as follows:
\begin{eqnarray}
S_{n,l} & = & g(h_{i}^{n} \cdot a_{i}^{l}) \nonumber
\end{eqnarray}
where $g(\cdot)$ is a scalar function.The $S_{n,l} \in R^{N \times L}$ measures how well the $n$-th row vector in $C_{i}$ semantically matches the $l$-th row vector in $A_{i}$.
After that, \emph{Text-to-Visual Attention} $\overline{h_{i}^{n}}$ and \emph{Visual-to-Text Attention} $\overline{a_{i}^{l}}$ will be calculated respectively according to the alignment matrix $S$.
The $\overline{h_{i}^{n}}$ calculation formula is as follows:
\begin{eqnarray}
w_{n}^{t2v}& =& {\rm softmax}(S_{n:}) \nonumber \\
\overline{h_{i}^{n}}& =& h^{n}_{i} + \sum\limits_{l}w_{nl}^{t2v}a_{i}^{n} \nonumber 
\end{eqnarray}
The $\overline{a_{i}^{l}}$ calculation formula is as follows:
\begin{eqnarray}
w_{l}^{v2t} &=& {\rm softmax}(S_{:l}) \nonumber \\
\overline{a_{i}^{l}} &=& a_{i}^{l} + \sum\limits_{i}w_{ln}^{v2t}h_{i}^{n} \nonumber 
\end{eqnarray}

Among them, $w_{n}^{t2v}$ signifies which image regions are most relevant to each source word. $w_{l}^{v2t}$ signifies which source words semantically match each visual region mostly. Thus, we can get the final visual feature maps $\overline{A}_{i}$=$(\overline{a}_{i}^{1},\overline{a}_{i}^{2},\ldots,\overline{a}_{i}^{L})$, and the vectors for the whole source sentence  $\overline{C}_{i}$=$(\overline{h}_{i}^{1},\overline{h}_{i}^{2},\ldots,\overline{h}_{i}^{N})$. 
Finally, we substituted $\overline{C}_{i}$ and $\overline{A}_{i}$ into formulas~(\ref{1}) and (\ref{2}) in Section~\ref{background} to obtain the time-dependent context vector $c_{t}$ and the time-dependent visual vector $i_{t}$. 


\section{Experiments}
\subsection{Data}
To evaluate our approach, we experimented with three commonly used machine translation dataset, including  multimodal machine translation dataset \emph{Multi30K}~\citep{elliott2016Multi30K} English-to-German (EN-DE), Global Voices English-to-German (EN-DE)\citep{GVoice}, and WMT' 16 (100k)
English-to-German (EN-DE).
\begin{description}
\item[\emph{Multi30K}] \emph{Multi30K} dataset consists of about 31k bilingual sentence-images pairs, . In this paper, we use 29K English to German parallel sentence pairs with visual annotations as the training set. The 1,014 English to German sentence pairs visual annotations are used as dev set. Finally, the test2016 test dataset is used for evaluation.

\item[Global Voices] Global Voices (EN-DE) dataset consists of more than 70k bilingual sentence pairs from summaries of news articles. We randomly sample 2000 data as dev set, 2000 as test set, and use the remained as training set.

\item[WMT'16 (100k)] WMT dataset (EN-DE) consists of more than 4.5M bilingual sentence pairs mainly from the proceedings of the European Parliament. In order to focus on evaluating the effectiveness of the retrieved visual information, we attempt to exclude the influence of data size, and randomly sampled 100k sentence pairs as our training set instead of the total 4.5M sentence pairs, which is similar to the number of sentences in the \emph{Multi30K} dataset and Global Voices.
We used Newstest2016 as the test set. 

\end{description}


\subsection{System Setting}
\label{subsec:setting}

\textbf{Image Retrieval Implementation}
We used the Microsoft Bing\footnote{\url{https://global.bing.com/images}} as image search engine. As described in Section \ref{image_load}, for each source language sentence, we build 5 search queries and collect 5 images for each sentence. Specifically, 
if the number of words is less than 5 after stopword filtering, we simplely repeat the keyword list several times to ensure that the number of remained words is enough for creating 5 search queries. 


\textbf{Model Implementation}:  
We implemented our proposed model on the top of \citet{su2021bi-co}, which was developed
based on OpenNMT \citep{klein2017opennmt}.
We used \emph{MOSES}\footnote{http://www.statmt.org/moses/} scripts to tokenize, normalize, and lowercase both source and target sentences.
For text encoder, we used bi-directional RNN with GRU to extract text features. One 256D single-layer RNN was used for both forward and backward.
For visual encoder, we used the $\mathrm{res4f}$ layer of pre-trained ResNet-50 \citep{he2016res50} to extract visual features.
We used Adam optimizer with mini-batches size of 32 to train all models, and set the learning rate as 0.001. 

We trained the model up to 15 epochs, and the training was early-stopped if BLEU \citep{papineni2002bleu} score of dev set did not improve for 3 epochs. The model with highest BLEU score of the dev set was selected to evaluate the test set. 
In order to reduce the influence of random seeds on the experimental results and ensure the stability of the final experimental results, 
we repeated the experiment 5 times with fixed 5 random seeds and used the macro average of BLEU scores as the final result.

\textbf{Baseline}
For each dataset, we used the text-only LSTM \citep{LSTM} as a baseline.

%
For \emph{Multi30K} dataset,  we quantitatively compared the proposed method with the following models:

\begin{itemize}
\item \citet{zhang2019UVR} used a text-only Transformer and proposed a universal visual representation method by retrieving images from a topic-image lookup table.
\item\citet{su2021bi-co} used a bi-direction attention network and a co-attention mechanism to enhance semantic interaction of text and images.
\item\citet{zhao2021word} proposed a novel integration strategy Word-Region Alignment(WRA) of the MNMT model that leverages the WRA to guide the model to translate certain source words into target words while attending to semantically relevant image regions.
\end{itemize}

We trained these models by employing the same training set and the same training parameters as the proposed model, and report the 4-gram BLEU score \citep{papineni2002bleu} for all baselines as well as the proposed method.

\begin{table}
    \centering
    \begin{tabular}{c c}
    \hline
    Method & BLEU Score \\ \hline \hline 
    \multicolumn{2}{c}{Text-only NMT} \\ \hline
    \!\!\!Bi-LSTM \citep{calixto2017doubly}\!\!\! & 33.7 \\
    \!\!\!Transformer \citep{zhang2019UVR}\!\!\! & 36.86 \\
   \hline \hline
    \multicolumn{2}{c}{Multimodal NMT with Original Images} \\ \hline
    \citet{zhang2019UVR} & 36.86 \\
    \citet{zhao2021word} & 38.40 \\ 
    \citet{su2021bi-co} & 39.20 \\
    \textcolor{black}{The proposed method} & 38.14 \\
    \hline \hline
    \multicolumn{2}{c}{Multimodal NMT with Retrieved Images} \\ \hline
    \citet{zhang2019UVR} & 36.94 \\
    The proposed method & \textbf{38.43} \\ \hline
    \end{tabular}
    \caption{Results on Multi30K}
    \label{tab:bleu_Multi30K}
\end{table}

\begin{table}
    \centering
    \vspace{0.5cm}
    \begin{tabular}{c c c}
    \hline
    System & \multicolumn{2}{c}{BLEU Score}  \\ \cline{2-3}
     & Global Voices & \!\!\!WMT'16 (100k)\!\!\! \\ \hline \hline
    Text-only  & \multirow{2}{*}{9.22} & \multirow{2}{*}{7.99} \\ 
    LSTM & & \\ \hline
    \!\!\!The proposed \!\!\! & \multirow{2}{*}{{\bf 9.81}} & \multirow{2}{*}{{\bf 8.41}} \\
    Method & & \\ \hline
    \end{tabular}
    \caption{Results on Global Voices and WMT`16 (100k)}
    \label{tab:other result}
\end{table}

    
    

\subsection{Experimental Results}
\label{result}

\begin{figure*}[t]
    \centering
    \includegraphics[scale=0.57]{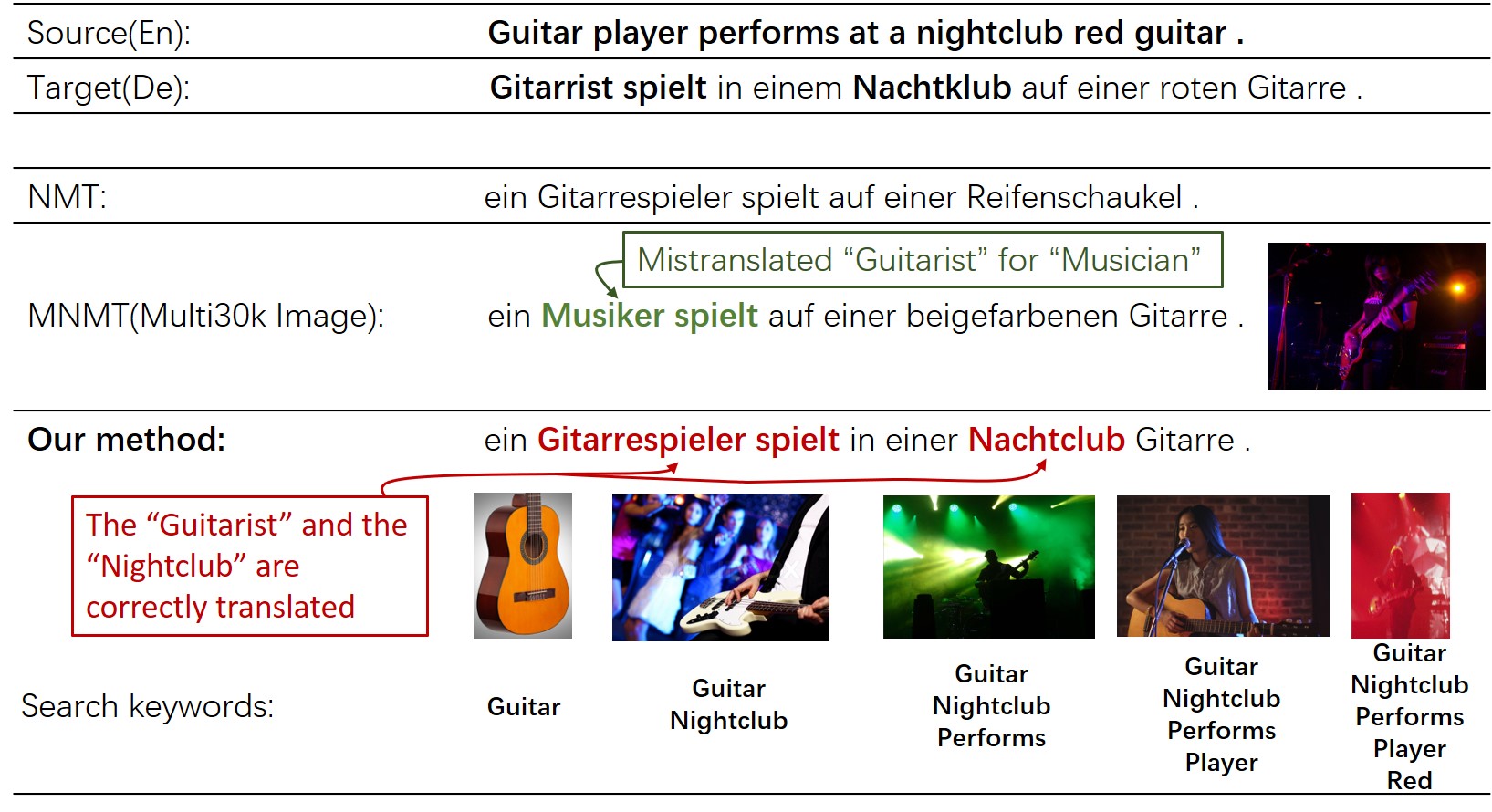}
    \caption{Example of correct translation by the proposed method}
    \label{fig:ex1}
\end{figure*}

Table \ref{tab:bleu_Multi30K} shows the experimental results on \emph{Multi30K} dataset. 
\textcolor{black}{The proposed method obtains a BLEU score of 38.43.}
Compared with the text-only NMT \citep{calixto2017doubly, vaswani2017transformer}, the proposed method obtains a significantly higher BLEU score. Compared with the multimodal NMT methods with original images \citep{zhang2019UVR,zhao2021word,su2021bi-co}, our proposed method obtains a comparable BLEU score\footnote{For \citet{su2021bi-co}, we trained the multimodal NMT model using the same parameters with our proposed method, and got a comprable BLEU score of 38.1 with our proposed method.}.
\textcolor{black}{Compared with the multimodal NMT method with retrieved images  \citep{zhang2019UVR}, the performance gain of the proposed method is approximately 1.5 BLEU.}

Futhermore, we quantitiatively compared our study with text-only NMT \citep{calixto2017doubly} on two dataset, i.e., Global Voices and WMT'16 (100k), which consist of bilingual sentence pairs without visual annotation. As shown in Table~\ref{tab:other result}, the proposed method achieved a higher BLUE score, demonstrating the effectiveness of the proposed search engine based image retrieval. More experimental results and discussions for the influence of collected images are described in Section~\ref{sec:anly}.





Figure~\ref{fig:ex1} shows an example of correct translation by the proposed method.
In this example, English words ``nightclub'' is failed to be translated by the model of \citet{su2021bi-co}, as well as the text-only NMT. It is mainly because that the text infomation is not enough for translating while the original image from \emph{Multi30K} is ambiguous and misleading. In the proposed method, we collected 5 images with image search engine according to the method described in Section~\ref{image_load}, among which 3 images provide effective visual information about ``nightclub'', and therefore, the proposed method correctly translate ``nightclub'' into ``Nachtclub''. Besides, benefit from visual information about ``guitar player'', the proposed method generates a partially correct translation ``Gitarrespieler spielt'', while is the model of \citet{su2021bi-co} incorrectly translate ``guitar player'' into ``Musiker spielt'' (musician).


\begin{table}[t]
    \centering
    \begin{tabular}{c c c}
    \hline
    Dataset & images & BLEU \\ \hline
    \multirow{4}{*}{Multi30K} & Text-only & 37.77 \\
    & Random Images & 37.65 \\ 
    & Blank Images & 37.79 \\ 
    & Retrieved Images & \textbf{38.43} \\ \hline
    
    \multirow{4}{*}{Global Voices}
    & Text-only & 9.22 \\ 
    & Random Images & 9.29 \\ 
    & Blank Images & 9.46 \\ 
    & Retrieved Images & \textbf{9.81} \\ \hline
    
    \multirow{4}{*}{\!\!\!WMT`16 (100k) \!\!\!}
    & Text-only & 7.99 \\ 
    & Random Images & 8.11 \\ 
    & Blank Images & 8.31 \\ 
    & Retrieved Images & \textbf{8.41} \\ \hline
    \end{tabular}
    \caption{Translation effect of different data sets under different image conditions}
    \label{tab:image compare}
\end{table}


\begin{figure}[t]
    \centering
    \includegraphics[scale=0.52]{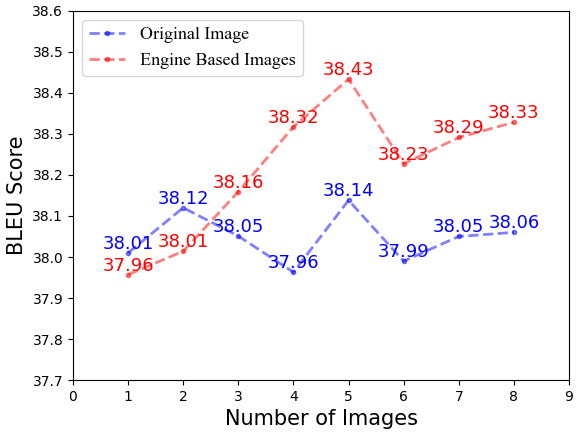}
    \caption{Influence of number of images on the BLEU score.}
    \label{fig:image nums}
\end{figure}

\section{Analysis and Discussion}
\label{sec:anly}

\begin{table*}[t]
    \centering
    \centering
    \begin{tabular}{c p{5cm} c}
    \hline
    Dataset & Sentence & Retrieved image \\ \hline \hline
    \emph{Multi30K} &  The person in the striped shirt is mountain climbing. & 
    \begin{tabular}{c}
    \specialrule{0em}{1pt}{1pt}
    \includegraphics[scale=0.35]{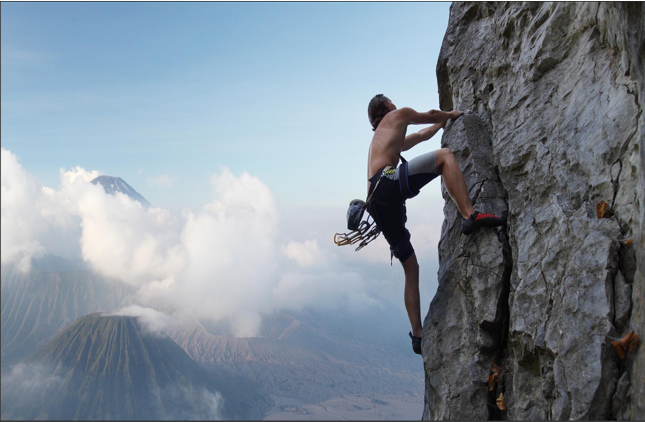}
    \end{tabular}
     \\ \hline
     \rule[-2pt]{0mm}{1.9cm}
    Global Voices & Now the city is under a siege from the security forces. &   
    \begin{tabular}{c}
    \includegraphics[scale=0.35]{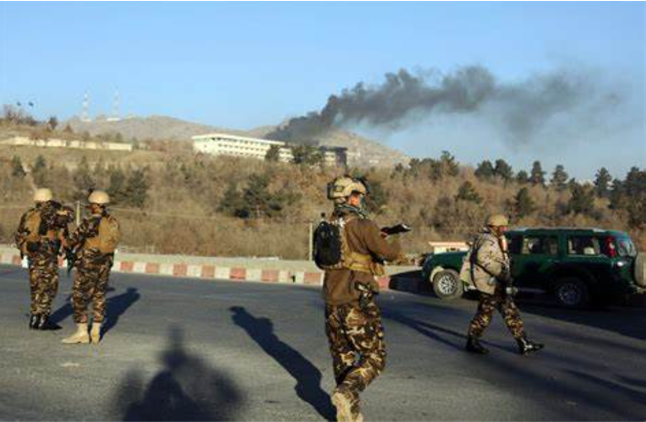}
    \end{tabular}
    \\ \hline
    \rule[-2pt]{0mm}{1.9cm}
    WMT'16 & In the future, integration will be a topic for the whole of society even more than it is today. &     \begin{tabular}{c}
    \includegraphics[scale=0.35]{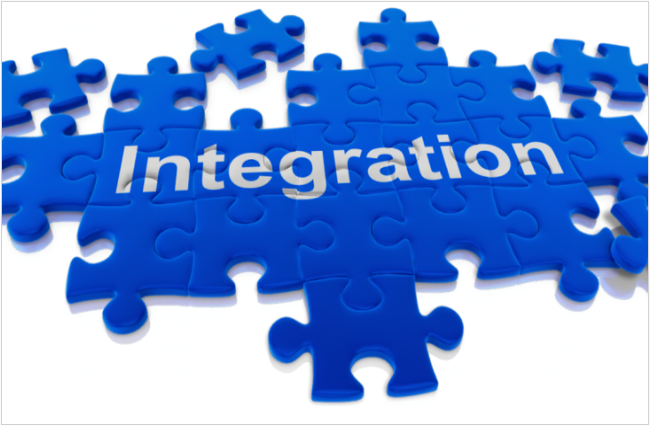}
    \end{tabular}
    \\ \hline
    \end{tabular}
    \caption{Examples of retrieved image from different datasets}
    \label{tab:image_exp}
\end{table*}

\subsection{Influence of the Number of Images}
\label{subsec:num_img}

For each sentence, several images can be obtained by following the image retrieval
method in section~\ref{image_load}.
To evaluate the influence of the number of paired images $m$, we constrained $m$ in $\{1,2,3,4,5,6,7,8\}$ for experiments on the \emph{Multi30K} dataset. As shown in figure~\ref{fig:image nums}, for different $m$, we used the images retrieved by search engine and the original images in \emph{Multi30K} dataset respectively for experiments. 
For images retrieved based on search engines,
as the number of images increases, the BLEU score also increased at the beginning(from 37.96 to 38.43) and then decreased when $m$ exceeds 5. The reason might be that retrieving too many images through search engines will lead to an increase in the number of noise images.
Therefore, we set $m=5$ in our models, and drawn a same conclusion as \citet{zhang2019UVR}.

For the original \emph{Multi30K} image, it only has the visual features of an image, so as the number of images increases, the BLEU score has no obvious upward trend.
In addition, when $m$ is less than 3, the BLEU score of the image using the original \emph{Multi30K} is higher than that of the retrieved image.

\subsection{Influence of the Quality of Images}
\label{subsec:qua_influ}

\begin{table}[t]
    \centering
    \begin{tabular}{c c}
    Dataset & Number of noise images  \\ \hline
    \emph{Multi30K} & 61  \\ 
    Global Voices & 228  \\
    WMT'16 & 685  \\ \hline
    \end{tabular}
    \caption{Number of noise images in 1000 collected images for each dataset}
    \label{tab:noise}
\end{table}

To evaluate the influence of the quality of collected images, we train the proposed model with randomly retrieved unrelated images, blank images, and retrieved images from image search engine, respectively. 
The evaluation results are shown in table~\ref{tab:image compare}.
It is obvious that proposed method achieves the highest BLEU score on all \emph{Multi30K} and Global Voices, demonstrating the effectiveness of visual information from collected images.

Compared with the model with random images and blank images, the performance gain of collected images is approximately 0.7 \& 0.6 BLEU score on \emph{Multi30K}, and 0.5 \& 0.3 BLUE score on Global Voices.
However, on the WMT'16 (100k) dataset, model with collected images obtains almost the same BLUE score as the model with blank images.

One of the possible reason is that sentences from WMT dataset contains fewer entity words that can be represented by images, and therefore, the proposed search engine based image retrieval method collects numbers of noise images.
Sentences from WMT'16 (100k) describe abstract concepts and complex events, while sentences from  
\emph{Multi30K} and Global Voices describe real objects and people, which is more reliable for image retrieval.
Examples of retrieved images of each dataset are shown in Table~\ref{tab:image_exp}. 
For the sentence from \emph{Multi30K} dataset, our method easily retrieves an image that represents ``A man is rock climbing''. For the sentence from Global Voice dataset, the retrieved image is partially consistent with the source sentence, containing contents of ``city'',``siege'' and``forces''. However, for the sentence from WMT'16 dataset, it is obvious that the retrieved image contains little effective visual information and can hardly provide assistance to translation.


%
%

To verify the hypotheses, we randomly sampled 1,000 images from the collected image set of each dataset, and manually classify the collected images into 2 classes, i.e., class of images that can provide visual information of the search query, and class of images that can not.
Images in second class are defined as noise images.
As shown in Table~\ref{tab:noise}, for \emph{Multi30K} dataset, only 61 out of 1000 collected images sampled are noise images, and the proportion is 6.1\%. 
%
However, in the WMT'16 dataset, the number of noise images obtained through retrieval is 685, accounting for more than half of the total number of images.Therefore, our method performs poorly on the WMT'16 dataset.
For the Global Voices dataset, the number of noise images is 228, which is between the \emph{Multi30K} and WMT'16 dataset, and the retrieved images also show better performance than the NMT model. 
It is insteresting to find that collected image set for \emph{Multi30K} has smallest proportion of noise image and achieves the biggest gain of translation performance, while the collected image set has the largest proportion of noise image and achieves the smallest gain of translation performance.


%
%

%

\section{Conclusions}
\label{sec:col}
In this paper, inspired by problem of \citet{zhang2019UVR} caused by applying limited collections of sentence-image pairs, 
we propose an open-vocabulary image retrieval methods to collect descriptive images for bilingual parallel corpus using image search engine, and introduce text-aware attentive visual encoder to filter incorrectly collected noise images.
%
%
Experiment results show that our proposed method achieves significant improvements over strong baselines, especially on \emph{Multi30K} and Global Voices. 
Further analysis shows that the effectiveness of the proposed methods in translating sentences that describe real objects and people.

\textcolor{black}{As one of our future work, we are going to evaluate our proposed method on some larger datasets, such as the entire WMT'16 dataset, and analyze the influence of the number of texts for the task of multimodal NMT.}
%
%

\bibliography{anthology,custom}
\bibliographystyle{acl_natbib}

\end{document}